\documentclass[10pt,twocolumn,letterpaper]{article}

\usepackage{arxiv}
\usepackage{times}
\usepackage{epsfig}
\usepackage{graphicx}
\usepackage{amsmath}
\usepackage{amssymb}

\usepackage{algorithm}
\usepackage{algpseudocode}
\algrenewcommand\algorithmicindent{.9em}%

\usepackage{comment}
\usepackage[british,UKenglish,USenglish,english,american]{babel}
\usepackage{multirow}
\usepackage{xcolor}
\usepackage{mathtools}

\usepackage{booktabs}
\usepackage{makecell}
\usepackage{subfigure}
\usepackage[pagebackref=true,breaklinks=true,letterpaper=true,colorlinks,bookmarks=false]{hyperref}

\begin{document}

\title{Grid R-CNN}

\author{Xin Lu$^1$\quad Buyu Li$^1$ \quad Yuxin Yue$^1$ \quad Quanquan Li$^1$ \quad Junjie Yan$^1$ \\
$^1$SenseTime Group Limited\\
{\tt\small \{luxin,libuyu,yueyuxin,liquanquan,yanjunjie\}@sensetime.com} \\
}
\maketitle

\begin{abstract}
This paper proposes a novel object detection framework named Grid R-CNN, which adopts a grid guided localization mechanism for accurate object detection. Different from the traditional regression based methods, the Grid R-CNN captures the spatial information explicitly and enjoys the position sensitive property of fully convolutional architecture. Instead of using only two independent points, we design a multi-point supervision formulation to encode more clues in order to reduce the impact of inaccurate prediction of specific points. To take the full advantage of the correlation of points in a grid, we propose a two-stage information fusion strategy to fuse feature maps of neighbor grid points. The grid guided localization approach is easy to be extended to different state-of-the-art detection frameworks. Grid R-CNN leads to high quality object localization, and experiments demonstrate that it achieves a 4.1\% AP gain at IoU=0.8 and a 10.0\% AP gain at IoU=0.9 on COCO benchmark compared to Faster R-CNN with Res50 backbone and FPN architecture.

\end{abstract}

\section{Introduction}
\label{sec:intro}

Object detection task can be decomposed into object classification and localization. In recent years, many deep convolutional neural networks (CNN) based detection frameworks are proposed and achieve state-of-the-art results~\cite{girshick2014rich,girshick2015fast,ren2015faster,lin2017feature,he2017mask,cai2017cascade}. Although these methods improve the detection performance in many different aspects, their bounding box localization modules are similar. Typical bounding box localization module is a regression branch, which is designed as several fully connected layers and takes in high-level feature maps to predict the offset of the candidate box (proposal or predefined anchor). 

\begin{figure}[t]
\centering
\includegraphics[width=1\linewidth]{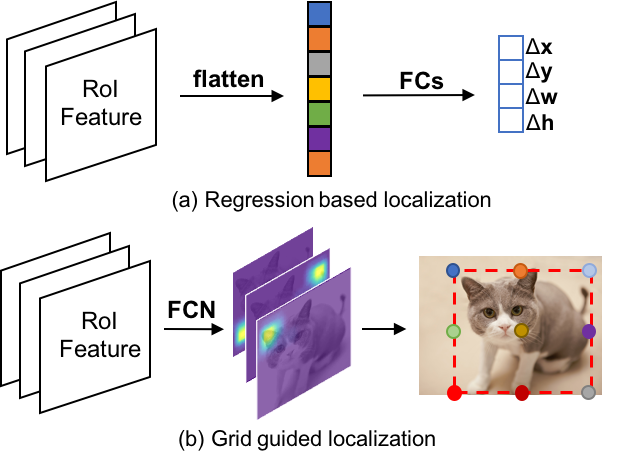}
\caption{(a) Traditional offset regression based bounding box localization. (b) Our proposed grid guided localization in Grid R-CNN. The bounding box is located by a fully convolutional network.}
\label{fig:intro} 
\end{figure}

In this paper we introduce Grid R-CNN, a novel object detection framework, where the traditional regression formulation is replaced by a grid point guided localization mechanism. And the explicit spatial representations are efficiently utilized for high quality localization.
In contrast to regression approach where the feature map is collapsed into a vector by fully connected layers, Grid R-CNN divides the object bounding box region into grids and employs a fully convolutional network (FCN)~\cite{long2015fully} to predict the locations of grid points. Owing to the position sensitive property of fully convolutional architecture, Grid R-CNN maintains the explicit spatial information and grid points locations can be obtained in pixel level.
As illustrated in Figure~\ref{fig:intro}.b, when a certain number of grid points at specified location are known, the corresponding bounding box is definitely determined. Guided by the grid points, Grid R-CNN can determine more accurate object bounding box than regression method which lacks the guidance of explicit spatial information.

Since a bounding box has four degrees of freedom, two independent points (e.g. the top left corner and bottom right corner) are enough for localization of a certain object. However the prediction is not easy because the location of the points are not directly corresponding to the local features. For example, the upper right corner point of the cat in Figure~\ref{fig:intro}.b lies outside of the object body and its neighborhood region in the image only contains background, and it may share very similar local features with nearby pixels. 
To overcome this problem, we design a multi-point supervision formulation. By defining target points in a gird, we have more clues to reduce the impact of inaccurate prediction of some points. For instance, in a typical $3 \times 3$ grid points supervision case, the probably inaccurate y-axis coordinate of the top-right point can be calibrated by that of top-middle point which just locates on the boundary of the object. The grid points are effective designs to decrease the overall deviation.

Furthermore, to take the full advantage of the correlation of points in a gird, we propose an information fusion approach. Specifically, we design individual group of feature maps for each grid point. For one grid point, the feature maps of the neighbor grid points are collected and fused into an integrated feature map. 
The integrated feature map is utilized for the location prediction of the corresponding grid point. Thus complementary information from spatial related grid points is incorporated to make the prediction more accurate.

We showcase the effectiveness of our Grid R-CNN framework on the object detection track of the challenging COCO benchmark \cite{lin2014microsoft}. Our approach outperforms traditional regression based state-of-the-art methods by a significant margin. For example, we surpass Faster R-CNN~\cite{ren2015faster} with a backbone of ResNet-50~\cite{he2016deep} and FPN~\cite{lin2017feature} architecture by 2.2\% AP. Further comparison on different IoU threshold criteria shows that our approach has overwhelming strength in high quality object localization, with a 4.1\% AP gain at IoU=0.8 and 10.0\% AP gain at IoU=0.9.

The main contributions of our work are listed as follows:
\begin{enumerate}
\item We propose a novel localization framework called Grid R-CNN which substitute traditional regression network by fully convolutional network that preserves spatial information efficiently. To our best knowledge, Grid R-CNN is the first proposed region based (two-stage) detection framework that locate object by predicting grid points on pixel level.

\item We design a multi-point supervision form that predicts points in grid to reduce the impact of some inaccurate points. We further propose a feature map level information fusion mechanism that enables the spatially related grid points to obtain incorporated features so that their locations can be well calibrated.

\item We perform extensive experiments and prove that Grid R-CNN framework is widely applicable across different detection frameworks and network architectures with consistent gains. The Grid R-CNN performs even better in more strict localization criterion (e.g. IoU threshold = 0.75). Thus we are confident that our grid guided localization mechanism is a better alternative for regression based localization methods.
\end{enumerate}

\begin{figure*}[ht]
\centering
\includegraphics[width=1\linewidth]{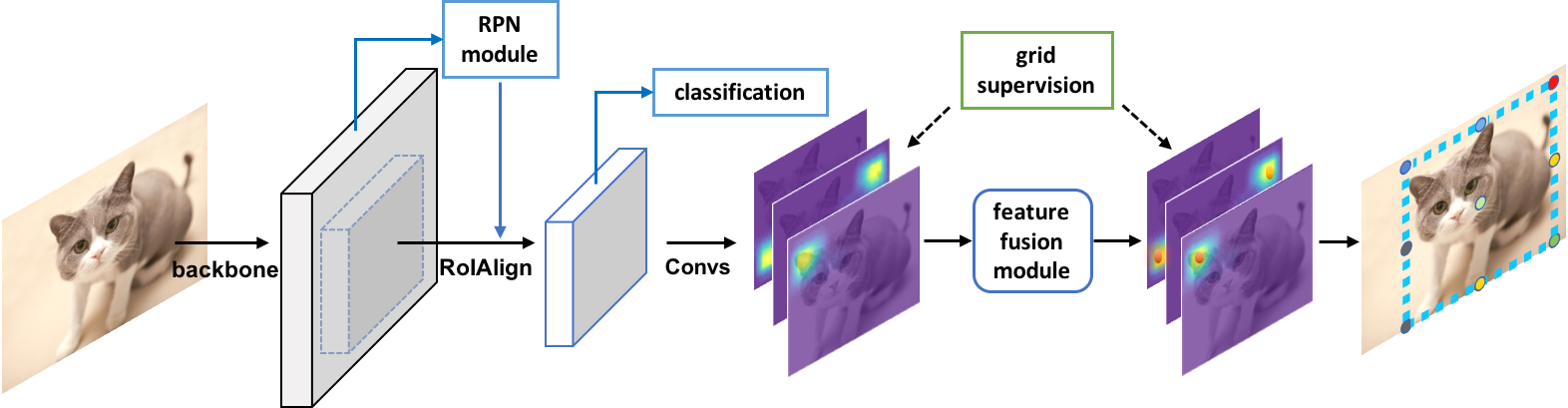} 
\caption{\textbf{Overview of the pipeline of Grid R-CNN.} Region proposals are obtained from RPN and used for RoI feature extraction from the output feature maps of a CNN backbone. The RoI features are then used to perform classification and localization. In contrast to previous works with a box offset regression branch, we adopt a grid guided mechanism for high quality localization. The grid prediction branch adopts a FCN to output a probability heatmap from which we can locate the grid points in the bounding box aligned with the object. With the grid points, we finally determine the accurate object bounding box by a feature map level information fusion approach.}
\label{fig:pipeline}
\end{figure*}

\section{Related Works}
Since our new approach is based on two stage object detector, here we briefly review some related works. Two-stage object detector was developed from the R-CNN architecture~\cite{girshick2014rich}, a region-based deep learning framework that classify and locate every RoI (Region of Interest) generated by some low-level computer vision algorithms~\cite{Uijlings2013Selective,zitnick2014edge}. Then SPP-Net~\cite{he2014spatial} and Fast-RCNN~\cite{girshick2015fast} introduced a new way to save redundant computation by extracting every region feature from the shared feature generated by entire image. Although SPP-Net and Fast-RCNN significantly improve the performance of object detection, the part of RoIs generating still cannot be trained end-to-end. Later, Faster-RCNN~\cite{ren2015faster} was proposed to solve this problem by utilizing a light region proposal network(RPN) to generate a sparse set of RoIs. This makes the whole detection pipeline an end-to-end trainable network and further improve the accuracy and speed of the detector. 

Recently, many works extend Faster R-CNN architecture in many aspects to achieve better performance.
For example, R-FCN~\cite{dai2016r} proposed to use region-based fully convolution network to replace the original fully connected network. FPN~\cite{lin2017feature} proposed a top-down architecture with lateral connections for building high-level semantic feature maps for variant scales. Mask R-CNN~\cite{he2017mask} extended Faster R-CNN by adding a branch for predicting an pixel-wise object mask in parallel with the original bounding box recognition branch. Different from Mask R-CNN which extends Faster R-CNN by adding a mask branch, our method replaces the regression branch with a new grid branch to locate objects more accurately. Also, our methods need no extra annotation other than bounding box.

CornerNet~\cite{law2018cornernet} is a single-stage object detector which uses paired key-points to locate the bounding box of the objects. It's a bottom-up detector that detects all the possible bounding box key-point(corner point) location through a hourglass~\cite{newell2016stacked} network. In the meanwhile, an embedding network was designed to map the paired keypoints as close as possible. With above embedding mechanism, detected corners can be group as pairs and locate the bounding boxes.

It's worth noting that our approach is quite different from CornerNet. CornerNet is a one-stage bottom-up method, which means it directly generate keypoints from the entire image without defining instance. So the key step of the CornerNet is to recognize which keypoints belong to the same instance and grouping them correctly. In contrast to that, our approach is a top-down two-stage detector which defines instance at first stage. What we focus on is how to locate the bounding box key-point more accurately. Furthermore, we designed grid points feature fusion module to exploit the features of related grid points and calibrate for more accurate grid points localization than two corner points only.

\section{Grid R-CNN}
An overview of Grid R-CNN framework is shown in Figure~\ref{fig:pipeline}. Based on region proposals, features for each RoI are extracted individually from the feature maps obtained by a CNN backbone. The RoI features are then used to perform classification and localization for the corresponding proposals. In contrast to previous works, e.g. Faster R-CNN, we use a grid guided mechanism for localization instead of offset regression. The grid prediction branch adopts a fully convolutional network~\cite{long2015fully}. It outputs a fine spatial layout (probability heatmap) from which we can locate the grid points of the bounding box aligned with the object. With the grid points, we finally determine the accurate object bounding box by a feature map level information fusion approach.


\subsection{Grid Guided Localization}
\label{sec:gl}
Most previous methods~\cite{girshick2014rich,girshick2015fast,ren2015faster,lin2017feature,he2017mask,cai2017cascade} use several fully connected layers as a regressor to predict the box offset for object localization. Whereas we adopt a fully convolutional network to predict the locations of predefined grid points and then utilize them to determine the accurate object bounding box.

We design an $N \times N$ grid form of target points aligned in the bounding box of object. An example of $3 \times 3$ case is shown in Figure~\ref{fig:intro}.b, the gird points here are the four corner points, midpoints of four edges and the center point respectively. Features of each proposal are extracted by RoIAlign~\cite{he2017mask} operation with a fixed spatial size of $14 \times 14$, followed by eight $3 \times 3$ dilated(for large receptive field) convolutional layers. After that, two 2$\times$ deconvolution layers are adopted to achieve a resolution of $56 \times 56$.
The grid prediction branch outputs $N \times N$ heatmaps with $56 \times 56$ resolution, and a pixel-wise sigmoid function is applied on each heatmap to obtain the probability map. And each heatmap has a corresponding supervision map, where 5 pixels in a cross shape are labeled as positive locations of the target grid point. Binary cross-entropy loss is utilized for optimization.

During inference, on each heatmap we select the pixel with highest confidence and calculate the corresponding location on the original image as the grid point. Formally, a point $(H_x, H_y)$ in heatmap will be mapped to the point $(I_x, I_y)$ in origin image by the following equation:
\begin{equation}
\begin{aligned}
\label{eq:mapping}
I_{x} &= P_{x}+\frac{H_x}{w_o}{w_p} \\ 
I_{y} &= P_{y}+\frac{H_y}{h_o}{h_p}
\end{aligned}
\end{equation}
where $(P_{x},P_{y})$ is the position of upper left corner of the proposal in input image, $w_p$ and $h_p$ are width and height of proposal, $w_o$ and $h_o$ are width and height of output heatmap. 

Then we determine the four boundaries of the box of object with the predicted grid points. Specifically, we denote the four boundary coordinates as $B = (x_l, y_u, x_r, y_b)$ representing the left, upper, right and bottom edge respectively. Let $g_j $ represent the j-th grid point with coordinate $(x_j, y_j)$ and predicted probability $p_j$,. Then we define $E_i$ as the set of indices of grid points that are located on the i-th edge, i.e., $j \in E_i$ if $g_j$ lies on the i-th edge of the bounding box.
We have the following equation to calculate $B$ with the set of $g$: 
\begin{equation}
\label{eq:boundary}
\begin{aligned}
x_l = \frac{1}{N}\sum_{j \in E_1} x_j p_j,
\qquad y_u = \frac{1}{N}\sum_{j \in E_2} y_j p_j \\
x_r = \frac{1}{N}\sum_{j \in E_3} x_j p_j,
\qquad y_b = \frac{1}{N}\sum_{j \in E_4} y_j p_j
\end{aligned}
\end{equation}
Taking the upper boundary $y_u$ as an example, it is the probability weighted average of y axis coordinates of the three upper grid points.

\begin{figure}[t]
\centering
\includegraphics[width=1\linewidth]{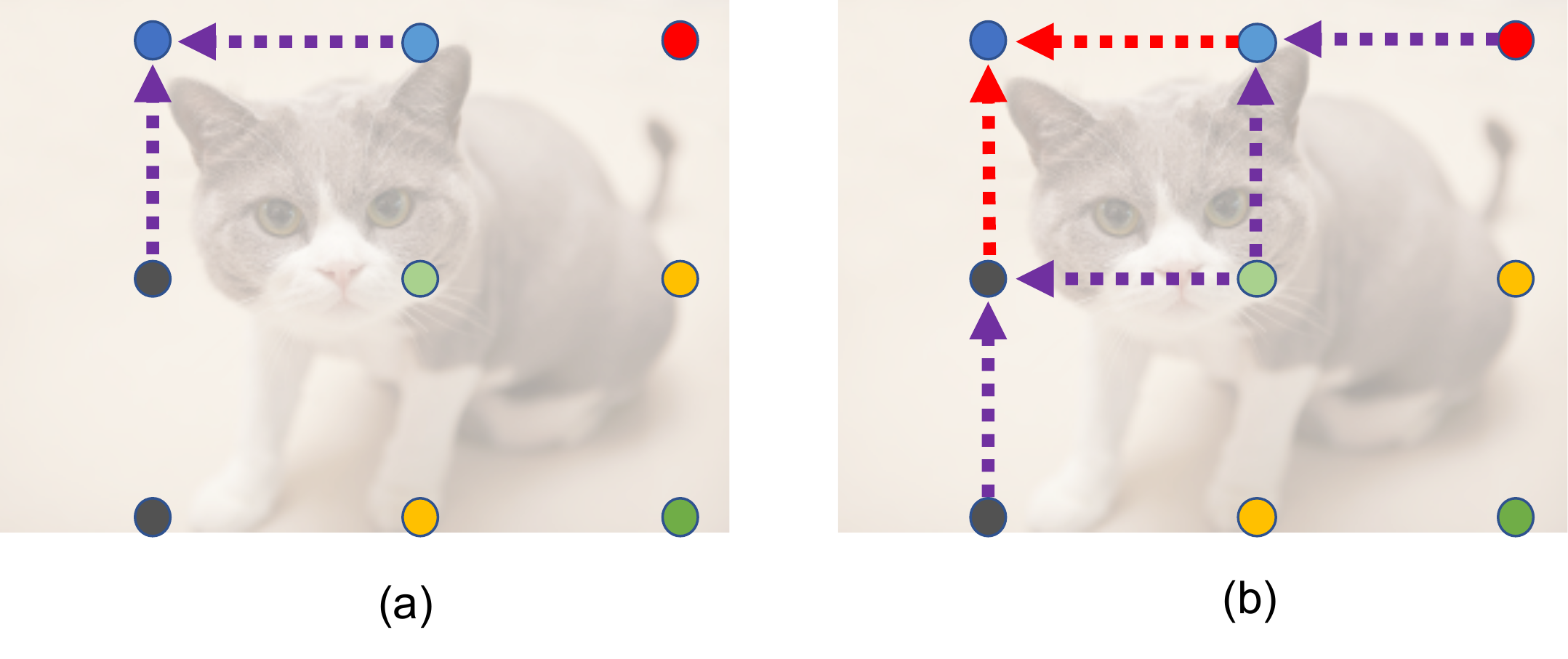}
\caption{An illustration of the $3 \times 3$ case of grid points feature fusion mechanism acting on the top left grid point. The arrows represent the spatial information transfer direction. (a) First order feature fusion, feature of the point can be enhanced by fusing features from its adjacent points. (b) The second order feature fusion design in Grid R-CNN.}
\label{fig:fusion}
\end{figure}

\subsection{Grid Points Feature Fusion}
The grid points have inner spatial correlation, and their locations can be calibrated by each other to reduce overall deviation. Thus a spatial information fusion module is designed.

An intuitive implementation is a coordinate level average, but the rich information in the feature maps are discarded. 
A further idea is to extract the local features corresponding to the grid points on each feature map for a fusion operation. However this also discards potential effective information in different feature maps. Taking the $3 \times 3$ gird as an example, for the calibration of top left point, the features in the top left region of other neighbor points' feature maps (e.g. the top middle point) may provide effective information but not used.
Therefore we design a feature map level information fusion mechanism to take full advantage of feature maps of each grid point. 

To distinguish the feature maps of different points, we use $N \times N$ group of filters to extract the features for them individually (from the last feature map) and give them intermediate supervision of their corresponding grid points. Thus each feature map has specified relationship with a certain grid point and we denote the feature map corresponding to the i-th point as $F_i$.

For each grid point, the points that have a $L_1$ distance of 1 (unit grid length) will contribute to the fusion, which are called source points. We define the set of source points w.r.t the i-th grid point as $S_i$. For the j-th source point in $S_i$, $F_j$ will be processed by three consecutive $5 \times 5$ convolution layers for information transfer and this process is denoted as a function $T_{j \rightarrow i}$. The processed features of all source points are then fused with $F_i$ to obtain an fusion feature map $F'_i$. An illustration of the top left grid point in $3 \times 3$ case is in Figure~\ref{fig:fusion}.a. We adopt a simple sum operation for the fusion in implementation and the information fusion is formulated as the following equation:
\begin{equation}
\label{eq:fusion}
F'_i = F_i + \sum_{j\in S_i } T_{j \rightarrow i}(F_j)
\end{equation}

Based on $F'_i$ for each grid point, a second order of fusion is then performed with new conv layers $T^+_{j \rightarrow i}$ that don't share parameters with those in first order of fusion. And the second order fused feature map $F''_i$ is utilized to output the final heatmap for the grid point location prediction. The second order fusion enables an information transfer in the range of 2 ($L_1$ distance). Taking the upper left grid point in $3 \times 3$ grids as an example (shown in Figure~\ref{fig:fusion}.b), it synthesizes the information from five other grid points for reliable calibration.

\begin{figure}[t]
\centering
\includegraphics[width=0.75\linewidth]{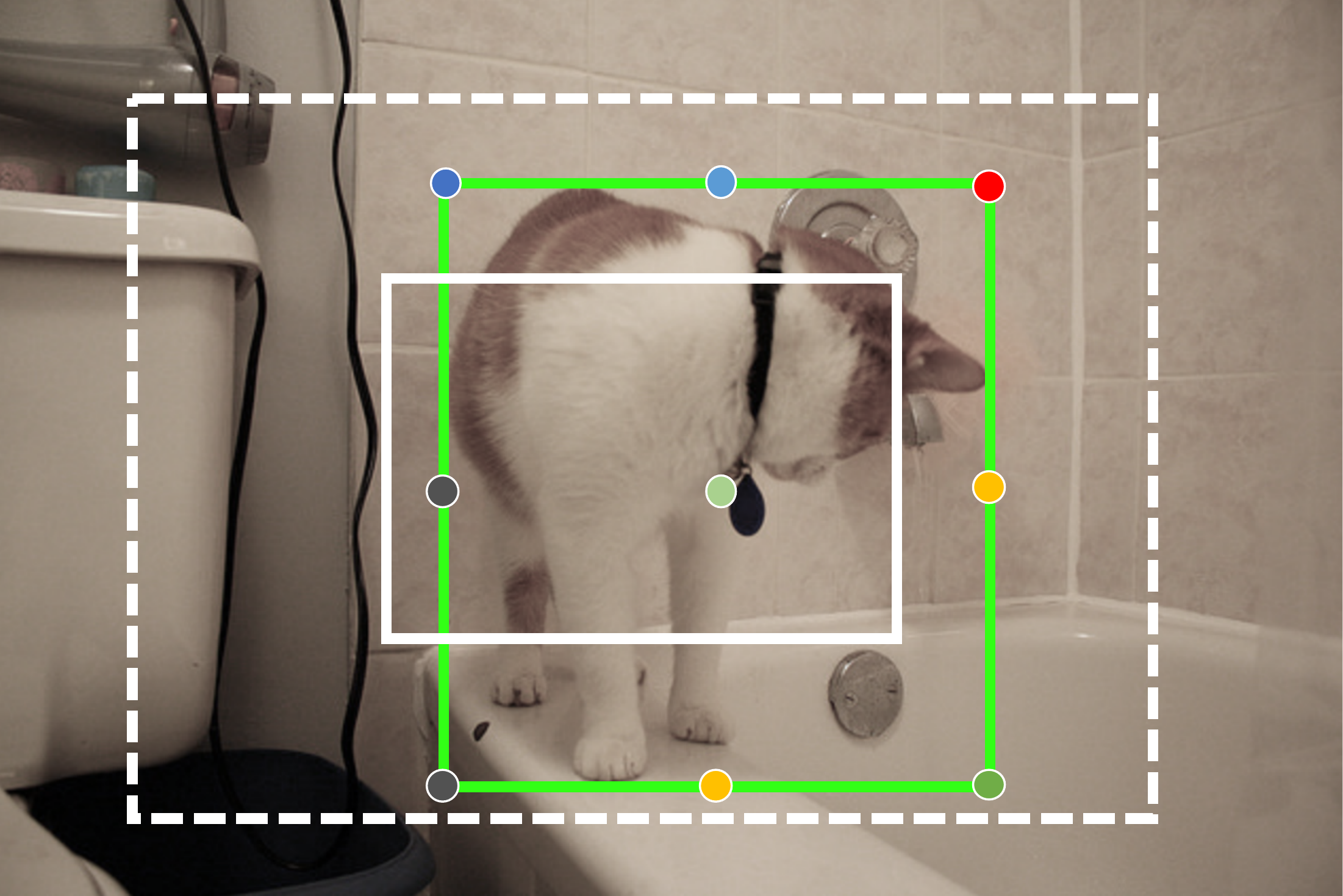}
\caption{Illustration of the extended region mapping strategy. The small white box is the original region of the RoI and we extend the representation region of the feature map to the dashed white box for higher coverage rate of the grid points in the the ground truth box which is in green.}
\label{fig:extend}
\end{figure}

\subsection{Extended Region Mapping}
\label{sec:cmapping}
Grid prediction module outputs heatmaps with a fixed spatial size representing the confidence distribution of the locations of grid points. Since the fully convolutional network architecture is adopted and spatial information is preserved all along, an output heatmap naturally corresponds to the spatial region of the input proposal in original image. However, a region proposal may not cover the entire object, which means some of the ground truth grid points may lie outside of the region of proposal and can't be labeled on the supervision map or predicted during inference. 

During training, the lack of some grid points labels leads to inefficient utilization of training samples. While in inference stage, by simply choosing the maximum pixel on the heatmap, we may obtain a completely incorrect location for the grid points whose ground truth location is outside the corresponding region. In many cases over half of the grid points are not covered, e.g. in Figure~\ref{fig:extend} the proposal (the small white box) is smaller than ground truth bounding box and 7 of the 9 grid points cannot be covered by output heatmap.

A natural idea is to enlarge the proposal area. This approach can make sure that most of the grid points will be included in proposal area, but it will also introduce redundant features of background or even other objects. Experiments show that simply enlarging the proposal area brings no gain but harms the accuracy of small objects detection.

To address this problem, we modify the relationship of output heatmaps and regions in the original image by a extended region mapping approach. Specifically, when the proposals are obtained, the RoI features are still extracted from the same region on the feature map without enlarging proposal area. While we re-define the representation area of the output heatmap as a twice larger corresponding region in the image, so that all grid points are covered in most cases as shown in Figure~\ref{fig:extend} (the dashed box).

The extended region mapping is formulated as a modification of Equation~\ref{eq:mapping}:
\begin{equation}
\begin{aligned}
I^{'}_{x} &= P_{x}+\frac{4H_x-w_o}{2w_o}{w_p} \\
I^{'}_{y} &= P_{y}+\frac{4H_y-h_o}{2h_o}{h_p}
\end{aligned}
\end{equation}
After the new mapping, all the target grid points of the positive proposals (which have an overlap larger than 0.5 with ground truth box) will be covered by the corresponding region of the heatmap.

\subsection{Implementation Details}
\textbf{Network Configuration}: We adopt the depth 50 or 101 ResNets \cite{he2016deep} w/o FPN~\cite{lin2017feature} constructed on top as backbone of the model. RPN~\cite{ren2015faster} is used to propose candidate regions. By convention, we set the shorter edge of the input image to 800 pixels in COCO dataset~\cite{lin2014microsoft} and 600 pixels in Pascal VOC dataset~\cite{everingham2015pascal}. In RPN, 256 anchors are sampled per image with 1:1 ratio of positive to negative anchors. The RPN anchors span 5 scales and 3 aspect ratios, and the IoU threshold of positive and negative anchors are 0.7 and 0.3 respectively. In classification branch, RoIs that have an overlap with ground truth greater than 0.5 are regarded as positive samples. We sample 128 RoIs per image in Faster R-CNN~\cite{ren2015faster} based model and 512 RoIs per image in FPN~\cite{lin2017feature} based model, with the 1:3 ratio of positive to negative. RoIAlign~\cite{he2017mask} is adopted in all experiments, and the pooling size is 7 in category classification branch and 14 in grid branch. The grid prediction branch samples at most 96 RoIs per image and only positive RoIs are sampled for training.

\textbf{Optimization}: We use SGD to optimize the training loss with 0.9 momentum and 0.0001 weight decay. The backbone parameter are initialized by image classification task on ImageNet dataset~\cite{russakovsky2015imagenet}, other new parameters are initialized by He (MSRA) initialization~\cite{HeZR015}. No data augmentations except standard horizontal flipping are used.
Our model is trained on 32 Nvidia TITAN Xp GPUs with one image on each for 20 epochs with an initial learning rate of 0.02, which decreases by 10 in the 13 and 18 epochs. We also use learning rate warming up and Synchronized BatchNorm machanism~\cite{goyal2017accurate,peng2017megdet} to make multi-GPU training more stable.

\textbf{Inference}: During the inference stage, the RPN generates 300/1000 (Faster R-CNN/FPN) RoIs per image. Then the features of these RoIs will be processed by RoIAlgin~\cite{he2017mask} layer and the classification branch to generate category score, followed by non-maximum suppression (NMS) with 0.5 IOU threshold. After that we select top 125 highest scoring RoIs and put their RoIAlign features into grid branch for further location prediction. Finally, NMS with 0.5 IoU threshold will be applied to remove duplicate detection boxes.


\section{Experiments}
We perform experiments on two object detection datasets, Pascal VOC~\cite{everingham2015pascal} and COCO~\cite{lin2014microsoft}. On Pascal VOC dataset, we train our model on VOC07+12 \textit{trainval} set and evaluate on VOC2007 \textit{test} set. On COCO~\cite{lin2014microsoft} dataset which contains 80 object categories, we train our model on the union of 80k \textit{train} images and 35k subset of \textit{val} images and test on a 5k subset of \textit{val} (\textit{minival}) and 20k \textit{test-dev}.



\subsection{Ablation Study}

\textbf{Multi-point Supervision}: Table~\ref{tab:density} shows how grid point selection affects the accuracy of detection. We perform experiments of variant grid formulations. The experiment of 2 points uses the supervision of upper left and bottom right corner of the ground truth box. In 4-point grid we add supervision of two other corner grid points. 9-point grid is a typical 3x3 grid formulation that has been described in section~\ref{sec:gl}. All experiments in Table~\ref{tab:density} are trained without feature fusion to avoid the extra gain from using more points for feature fusion. It can be observed that as the number of supervised grid points increases, the accuracy of the detection also increases.
\begin{table}[ht]
\begin{center}
\begin{tabular}{ l | c | c  c }
\hline
method & AP & $\text{AP}_{.5}$ & $\text{AP}_{.75}$ \\
\hline
regression  & 37.4 & 59.3 & 40.3 \\
2 points  & 38.3 & 57.3 & 40.5 \\
4-point grid  & 38.5 & 57.5 & 40.8 \\
9-point grid  & 38.9 & 58.2 & 41.2 \\
\hline
\end{tabular}
\vspace{1mm}
\caption{Comparison of different grid points strategies in Grid R-CNN. Experiments show that more grid points bring performance gains.}
\label{tab:density}
\vspace{-5mm}
\end{center}
\end{table}

\textbf{Grid Points Feature Fusion}: Results in Table~\ref{tab:fusion} shows the effectiveness of feature fusion. We perform experiments on several typical feature fusion methods and achieve different levels of improvement on AP performance. The bi-directional fusion method, as mentioned in~\cite{chu2016structured}, models the information flow as a bi-directional tree. 
For fair comparison, we directly use the feature maps from the first order feature fusion stage for grid point location prediction, and see a same gain of 0.3\% AP as bi-directional fusion. And we also perform experiment of the complete two stage feature fusion. As can be seen in Table~\ref{tab:fusion}, the second order fusion further improves the AP by 0.4\%, with a 0.7\% gain from the non-fusion baseline. Especially, the improvement of $AP_{0.75}$ is more significant than that of $AP_{0.5}$, which indicates that feature fusion mechanism helps to improve the localization accuracy of the bounding box.

\begin{table}[t]
\begin{center}
\begin{tabular}{ l | c | c  c }
\hline
method & AP & $\text{AP}_{.5}$ & $\text{AP}_{.75}$ \\
\hline
w/o fusion  & 38.9 & 58.2 & 41.2 \\
bi-directional fusion~\cite{chu2016structured}  & 39.2 & 58.2 & 41.8  \\
first order feature fusion  & 39.2 & 58.1 & 41.9  \\
second order feature fusion  & 39.6 & 58.3 & 42.4 \\
\hline
\end{tabular}
\vspace{1mm}
\caption{Comparison of different feature fusion methods. Bi-directional feature fusion, first order feature fusion and second order fusion all demonstrate improvements. Second order fusion achieves the best performance with an improvement of 0.7\% on AP.}
\label{tab:fusion}
\end{center}
\end{table}

\textbf{Extended Region Mapping}: Table~\ref{tab:extend} shows the results of our extended region mapping strategy compared with the original region representation and the method of directly enlarging the proposal box. Directly enlarging the region of proposal box for RoI feature extraction helps to cover more grid points of big objects but also brings in redundant information for small objects. Thus we can see that with this enlargement method there is a increase in $AP_{large}$ but a decrease in $AP_{small}$, and finally a decline compared with the baseline. Whereas the extended region mapping strategy improves $AP_{large}$ performance as well as producing no negative influences on $AP_{small}$, which leads to 1.2\% improvement on AP. 
\begin{table}[t]
\begin{center}
\begin{tabular}{ l | c | c  c }
\hline
method & AP & $\text{AP}_{small}$ & $\text{AP}_{large}$ \\
\hline
baseline  & 37.7 & 22.1 & 48.0 \\
enlarge proposal area  & 37.7 & 20.8 & 50.9  \\
extended region mapping  & 38.9 & 22.1 & 51.4  \\
\hline
\end{tabular}
\vspace{1mm}
\caption{Comparison of enlarging the proposal directly and extended region mapping strategy.}
\label{tab:extend}
\vspace{-5mm}
\end{center}
\end{table}

\subsection{Comparison with State-of-the-art Methods}
On \textit{minival} set, we mainly compare Grid R-CNN with two widely used two-stage detectors, Faster-RCNN and FPN. We replace the original regression based localization method by the grid guided localization mechanism in the two frameworks for fair comparison. 

\textbf{Experiments on Pascal VOC}: We train Grid R-CNN on Pascal VOC dataset for 18 epochs with the learning rate reduced by 10 at 15 and 17 epochs. The origianl evaluation criterion of PASCAL VOC is to calculate the mAP at 0.5 IoU threshold. We extend that to the COCO-style criterion which calculates the average AP across IoU thresholds from 0.5 to 0.95 with an interval of 0.05. We compare Grid R-CNN with R-FCN~\cite{dai2016r} and FPN~\cite{lin2017feature}. Results in Table~\ref{tab:voc} show that our Grid R-CNN significantly improve AP over FPN and R-FCN by 3.6\% and 9.7\% respectively.
\begin{table}[t]
\begin{center}
\begin{tabular}{ l | l | c  }
\hline
method & backbone & AP \\
\hline
R-FCN & ResNet-50  & 45.6  \\
FPN & ResNet-50  & 51.7  \\
FPN based Grid R-CNN & ResNet-50  & 55.3 \\
\hline
\end{tabular}
\vspace{1mm}
\caption{Comparison with R-FCN and FPN on Pascal VOC dataset. Note that we evaluate the results with a COCO-style criterion which is the average AP across IoU thresholds range from 0.5 to [0.5:0.95].}
\label{tab:voc}
\vspace{-5mm}
\end{center}
\end{table}

\textbf{Experiments on COCO}: To further demonstrate the generalization capacity of our approach, we conduct experiments on challenging COCO dataset. Table~\ref{tab:coco} shows that our approach brings consistently and substantially improvement across multiple backbones and frameworks. Compared with Faster R-CNN framework, Grid R-CNN improves AP over baseline by 2.1\% with ResNet-50 backbone. The significant improvements are also shown on FPN framework based on both ResNet-50 and ResNet-101 backbones. Experiments in Table~\ref{tab:coco} demonstrate that Grid R-CNN significantly improve the performance of middle and large objects by about 3 points.
\begin{table*}[t]
\begin{center}
\begin{tabular}{ l | c | c | c  c | c  c  c }
\hline
method & backbone & AP & $\text{AP}_{.5}$ & $\text{AP}_{.75}$ & $\text{AP}_{S}$ & $\text{AP}_{M}$ & $\text{AP}_{L}$ \\
\hline
Faster R-CNN & ResNet-50 & 33.8 & 55.4 & 35.9 & 17.4 & 37.9 & 45.3 \\
Grid R-CNN & ResNet-50 & \textbf{35.9} & 54.0 & 38.0 & 18.6 & 40.2 & 47.8 \\
\hline
Faster R-CNN w FPN & ResNet-50 & 37.4 & 59.3 & 40.3 & 21.8 & 40.9 & 47.9 \\
Grid R-CNN w FPN & ResNet-50 & \textbf{39.6} & 58.3 & 42.4 & 22.6 & 43.8 & 51.5 \\
\hline
Faster R-CNN w FPN & ResNet-101 & 39.5 & 61.2 & 43.1 & 22.7 & 43.7 & 50.8 \\
Grid R-CNN w FPN & ResNet-101 & \textbf{41.3} & 60.3 & 44.4 & 23.4 & 45.8 & 54.1 \\
\hline
\end{tabular}
\vspace{1mm}
\caption{Bounding box detection AP on COCO \textit{minival}. Grid R-CNN outperforms both Faster R-CNN and FPN on ResNet-50 and ResNet-101 backbone. }
\label{tab:coco}
\end{center}
\end{table*}

\textbf{Results on COCO \textit{test-dev} Set}: For complete comparison, we also evaluate Grid R-CNN on the COCO \textit{test-dev} set. We adopt ResNet-101 and ResNeXt-101~\cite{xie2017aggregated} with FPN~\cite{lin2017feature} constructed on the top. Without bells and whistles, Grid R-CNN based on ResNet-101-FPN and ResNeXt-101-FPN could achieve 41.5 and 43.2 AP respectively. As shown in Table~\ref{tab:test}, Grid R-CNN achieves very competitive performance comparing with other state-of-the-art detectors. It outperforms Mask R-CNN by a large margin without using any extra annotations. Note that since the techniques such as scaling used in SNIP~\cite{DBLP:journals/corr/abs-1711-08189} and cascading in Cascade R-CNN~\cite{cai2017cascade} are not applied in current framework of Grid R-CNN, there is still room for large improvement on performance (e.g. combined with scaling and cascading methods).

\begin{table*}[t]
\begin{center}
\begin{tabular}{ l | l | c | c  c | c  c  c }
\hline
method & backbone & AP & $\text{AP}_{.5}$ & $\text{AP}_{.75}$ & $\text{AP}_{S}$ & $\text{AP}_{M}$ & $\text{AP}_{L}$ \\
\hline
YOLOv2~\cite{redmon2016yolo9000} & DarkNet-19  & 21.6 & 44.0 & 19.2 & 5.0 & 22.4 & 35.5 \\
SSD-513~\cite{DBLP:journals/corr/LiuAESR15} & ResNet-101  & 31.2 & 50.4 & 33.3 & 10.2 & 34.5 & 49.8 \\
DSSD-513~\cite{fu2017dssd} & ResNet-101  & 33.2 & 53.3 & 35.2 & 13.0 & 35.4 & 51.1 \\
RefineDet512~\cite{zhang2017single} & ResNet101  & 36.4 & 57.5 & 39.5 & 16.6 & 39.9 & 51.4 \\
RetinaNet800~\cite{lin2018focal} & ResNet-101  & 39.1 & 59.1 & 42.3 & 21.8 & 42.7 & 50.2 \\
CornerNet & Hourglass-104  & 40.5 & 56.5 & 43.1 & 19.4 & 42.7 & 53.9 \\
\hline
Faster R-CNN+++~\cite{he2016deep} & ResNet-101  & 34.9 & 55.7 & 37.4 & 15.6 & 38.7 & 50.9 \\
Faster R-CNN w FPN~\cite{lin2017feature} & ResNet-101  & 36.2 & 59.1 & 39.0 & 18.2 & 39.0 & 48.2 \\
Faster R-CNN w TDM~\cite{shrivastava2016beyond} & Inception-ResNet-v2~\cite{szegedy2017inception}  & 36.8 & 57.7 & 39.2 & 16.2 & 39.8 & 52.1 \\
D-FCN~\cite{dai2017deformable} & Aligned-Inception-ResNet & 37.5 & 58.0 & - & 19.4 & 40.1 & 52.5 \\
Regionlets~\cite{xu2017deep} & ResNet-101 & 39.3 & 59.8 & - & 21.7 & 43.7 & 50.9 \\
Mask R-CNN~\cite{he2017mask}  & ResNeXt-101 & 39.8 & 62.3 & 43.4 & 22.1 & 43.2 & 51.2 \\
Grid R-CNN w FPN (ours) & ResNet-101 & 41.5 & 60.9 & 44.5 & 23.3 & 44.9 & 53.1 \\
Grid R-CNN w FPN (ours) & ResNeXt-101  & 43.2 & 63.0 & 46.6 & 25.1 & 46.5 & 55.2 \\
\end{tabular}
\vspace{1mm}
\caption{Comparison with state-of-the-art detectors on COCO \textit{test-dev}.}
\label{tab:test}
\end{center}
\end{table*}

\subsection{Analysis and Discussion}

\begin{figure}[h] 
\centering 
\includegraphics[width=1\linewidth]{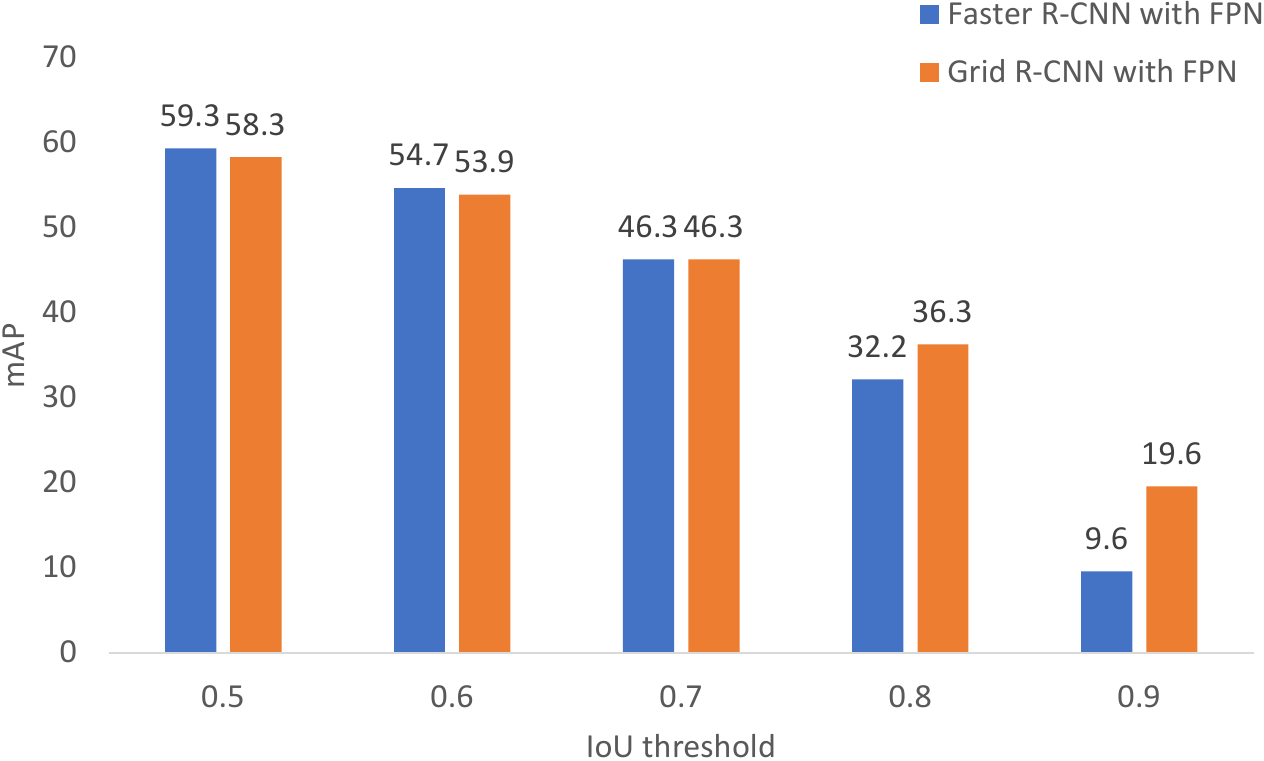}
\caption{AP results across IoU thresholds from 0.5 to 0.9 with an interval of 0.1.} 
\label{Fig.hist} 
\end{figure}

\textbf{Accuracy in Different IoU Criteria}: In addition to the overview of mAP, in this part we focus on the localization quality of the Grid R-CNN.
Figure~\ref{Fig.hist} shows the comparison between FPN based Grid R-CNN and baseline FPN with the same ResNet-50 backbone across IoU thresholds from 0.5 to 0.9. Grid R-CNN outperforms regression at higher IoU thresholds (greater than 0.7). The improvements over baseline at AP$_{0.8}$ and AP$_{0.9}$ are 4.1\% and 10\% respectively, which means that Grid R-CNN achieves better performance mainly by improving the localization quality of the bounding box. In addition, the results of AP$_{0.5}$ indicates that grid branch may slightly affect the performance of the classification branch.

\begin{figure*}[t]
\centering
\label{qualitative}
\includegraphics[width=1\linewidth]{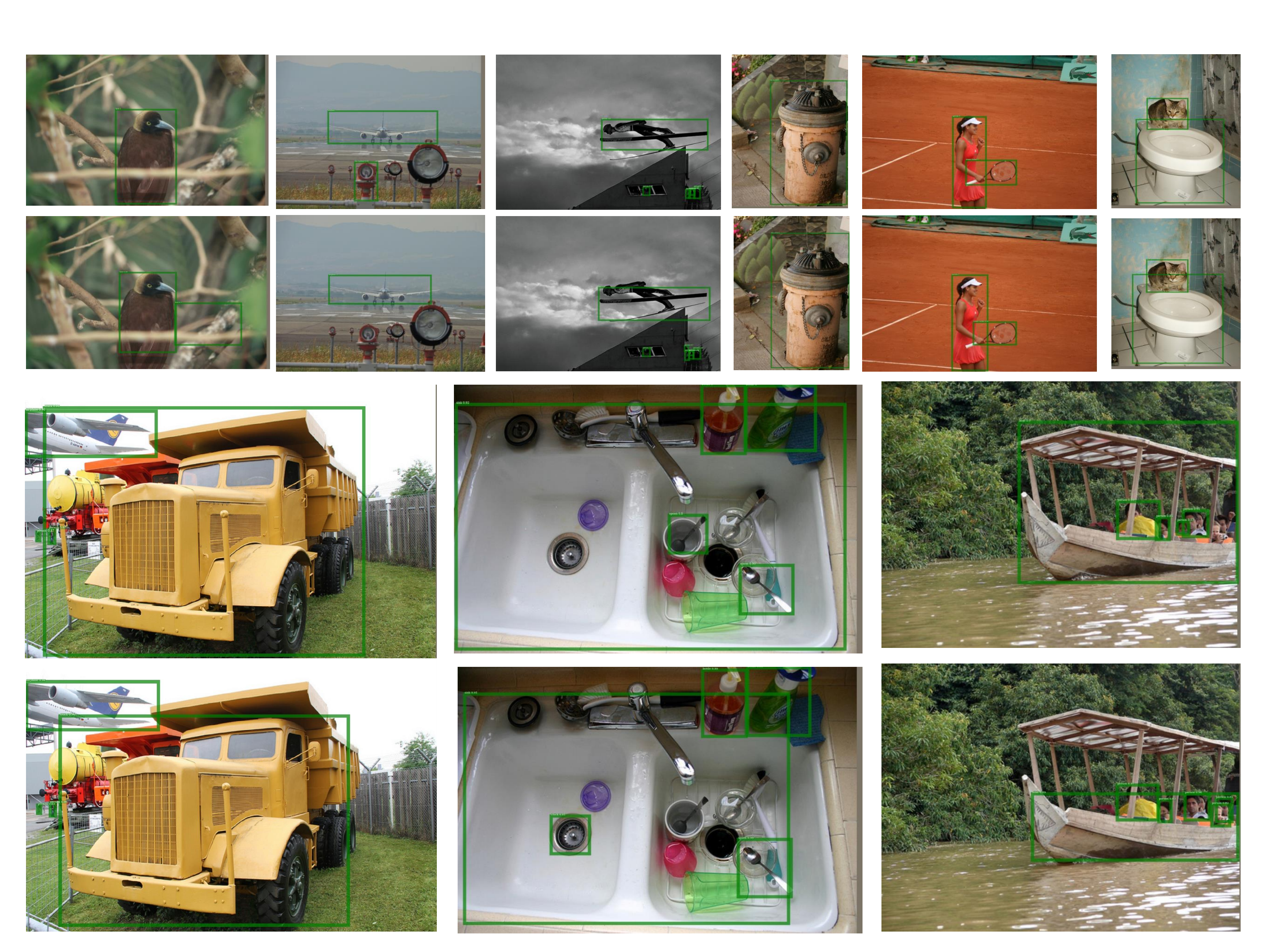}
\caption{Qualitative results comparison. The results of Grid R-CNN are listed in the first and third row, while those of Faster R-CNN are in the second and fourth row. }
\label{fig:qualitative}
\end{figure*}

\begin{table*}[ht]
\setlength{\tabcolsep}{0.5mm}
\footnotesize
\begin{center}
\begin{tabular}{ | c | c | c | c  | c | c | c | c | c | c | c  |c | c | c | c | c  |  }
\hline
category & cat & bear & giraffe & dog & airplane & horse & zebra & toilet & keyboard & fork & teddy bear & train & laptop & refrigerator & hot dog\\
\hline
gain & 6.0 & 5.6 & 5.4 & 5.3 & 5.3 & 5.0 & 4.8 & 4.8 & 4.7 &  4.6 & 4.4 & 4.2 & 4.0 & 3.6 & 3.6\\
\hline
category & toaster & hair drier & sports ball & frisbee& traffic light & backpack & kite & handbag & microwave & bowl & clock & cup & carrot & dining table & boat\\
\hline
gain & -1.9& -1.3 & -1.0 & -0.8 & -0.5 & -0.4 & -0.3 & -0.1 & -0.1 & -0.1 & 0.1 & 0.1 & 0.2 & 0.3 & 0.3 \\
\hline
\end{tabular}
\vspace{1mm}
\caption{The top 15 categories with most gains and most declines respectively, in the results of Grid R-CNN compared to Faster R-CNN.}
\label{tab:top10}
\end{center}
\end{table*}

\textbf{Varying Degrees of Improvement in Different Categories}: We have analyzed the specific improvement of Grid R-CNN on each category and discovered a meaningful and interesting phenomenon. As shown in Table~\ref{tab:top10}, the categories with the most gains usually have a rectangular or bar like shape (e.g. keyboard, laptop, fork, train, and refrigerator), while the categories suffering declines or having least gains usually have a round shape without structural edges (e.g. sports ball, frisbee, bowl, clock and cup). This phenomenon is reasonable since grid points are distributed in a rectangular shape. Thus the rectangular objects tend to have more grid points on the body but round objects can never cover all the grid points (especially the corners) with its body. Moreover, we are inspired to design points in circle shapes for better localization of objects with a round shape in future works.

\textbf{Qualitative Results Comparison}: We showcase the illustrations of our high quality object localization results in this part. As shown in Figure~\ref{fig:qualitative}, Grid R-CNN (in the 1st and 3rd row) has an outstanding performance in accurate localization compared with the widely used Faster R-CNN (in the 2nd and 4th row). First and second row in figure~\ref{fig:qualitative} show that Grid R-CNN outperforms Faster R-CNN in high quality object detection. Third and 4th row show that Grid R-CNN performs better in large object detection tasks.

\section{Conclusion}
In this paper we propose a novel object detection framework, Grid R-CNN, which replaces the traditional box offset regression strategy in object detection by a grid guided mechanism for high quality localization. The grid branch locates the object by predicting grid points with the position sensitive merits of FCN and then determining the bounding box guided by the grid. Further more, we design a feature fusion module to calibrate the locations of grid points by transferring the spatial information in feature map level. Additionally, an extended region mapping mechanism is proposed to help RoIs get a larger representing area to cover as many grid points as possible, which significantly improves the performance. 
Extensive experiments show that Grid R-CNN brings solid and consistent improvement and achieves state-of-the-art performance, especially on strict evaluation metrics such as AP at IoU=0.8 and IoU=0.9. Since the grid guided localization approach is easy to be extended to other frameworks, we will try to combine the scale selection and cascade techniques with Grid R-CNN and we believe a further gain can be obtained.

{\small
\bibliographystyle{ieee}
\bibliography{gridbib}

\begin{thebibliography}{10}\itemsep=-1pt

\bibitem{girshick2014rich}
R.~Girshick, J.~Donahue, T.~Darrell, and J.~Malik.
\newblock Rich feature hierarchies for accurate object detection and semantic
  segmentation.
\newblock In {\em CVPR}, 2014.

\bibitem{girshick2015fast}
R.~Girshick.
\newblock {Fast R-CNN}.
\newblock In {\em ICCV}, 2015.

\bibitem{ren2015faster}
S.~Ren, K.~He, R.~Girshick, and J.~Sun.
\newblock {Faster R-CNN: Towards real-time object detection with region
  proposal networks}.
\newblock In {\em NIPS}, 2015.

\bibitem{lin2017feature}
T.-Y. Lin, P.~Doll{\'a}r, R.~Girshick, K.~He, B.~Hariharan, and S.~Belongie.
\newblock Feature pyramid networks for object detection.
\newblock In {\em CVPR}, 2017.

\bibitem{he2017mask}
K.~He, G.~Gkioxari, P.~Doll{\'a}r, and R.~Girshick.
\newblock Mask r-cnn.
\newblock In {\em ICCV}, 2017.

\bibitem{cai2017cascade}
Cai, Z., Vasconcelos, N.: Cascade r-cnn: Delving into high quality object
  detection. arXiv preprint arXiv:1712.00726  (2017)

\bibitem{long2015fully}
Long, Jonathan and Shelhamer, Evan and Darrell, Trevor.
\newblock {Fully convolutional networks for semantic segmentation}.
\newblock In {\em CVPR}, 2015.

\bibitem{he2016deep}
K.~He, X.~Zhang, S.~Ren, and J.~Sun.
\newblock Deep residual learning for image recognition.
\newblock In {\em CVPR}, 2016.

\bibitem{law2018cornernet}
Law, Hei and Deng, Jia.
\newblock {Cornernet: Detecting objects as paired keypoints}
\newblock In {\em ECCV}, 2018.

\bibitem{lin2014microsoft}
T.-Y. Lin, M.~Maire, S.~Belongie, J.~Hays, P.~Perona, D.~Ramanan,
  P.~Doll{\'a}r, and C.~L. Zitnick.
\newblock Microsoft coco: Common objects in context.
\newblock In {\em European conference on computer vision}, pages 740--755.
  Springer, 2014.

\bibitem{he2014spatial}
K.~He, X.~Zhang, S.~Ren, and J.~Sun.
\newblock Spatial pyramid pooling in deep convolutional networks for visual
  recognition.
\newblock In {\em ECCV}, pages 346--361, 2014.

\bibitem{dai2016r}
Dai, J., Li, Y., He, K., Sun, J.: R-fcn: Object detection via region-based
  fully convolutional networks. arXiv preprint arXiv:1605.06409  (2016)

\bibitem{newell2016stacked}
Newell, A., Yang, K., Deng, J.: Stacked hourglass networks for human pose
  estimation. In: European Conference on Computer Vision. pp. 483--499.
  Springer (2016)

\bibitem{redmon2016yolo9000}
Redmon, J., Farhadi, A.: Yolo9000: better, faster, stronger. arXiv preprint
  1612 (2016)

\bibitem{DBLP:journals/corr/LiuAESR15}
W.~Liu, D.~Anguelov, D.~Erhan, C.~Szegedy, S.~E. Reed, C.~Fu, and A.~C. Berg.
\newblock {SSD:} single shot multibox detector.
\newblock In {\em ECCV}, pages 21--37, 2016.

\bibitem{fu2017dssd}
Fu, C.Y., Liu, W., Ranga, A., Tyagi, A., Berg, A.C.: Dssd: Deconvolutional
  single shot detector. arXiv preprint arXiv:1701.06659  (2017)

\bibitem{zhang2017single}
Zhang, Shifeng and Wen, Longyin and Bian, Xiao and Lei, Zhen and Li, Stan Z.: Single-shot
 refinement neural network for object detection. arXiv preprint arXiv:1711.06897 (2017)

\bibitem{lin2018focal}
T.-Y. Lin, P.~Goyal, R.~Girshick, K.~He, and P.~Doll{\'a}r.
\newblock Focal loss for dense object detection.
\newblock In {\em ICCV}, 2017.

\bibitem{shrivastava2016beyond}
Shrivastava, Abhinav and Sukthankar, Rahul and Malik, Jitendra and Gupta, Abhinav: Beyond
 skip connections: Top-down modulation for object detection. arXiv preprint arXiv:1612.06851 (2016)

\bibitem{dai2017deformable}
J.~Dai, H.~Qi, Y.~Xiong, Y.~Li, G.~Zhang, H.~Hu, and Y.~Wei.
\newblock Deformable convolutional networks.
\newblock In {\em ICCV}, 2017.

\bibitem{xu2017deep}
Xu, Hongyu and Lv, Xutao and Wang, Xiaoyu and Ren, Zhou and Chellappa, Rama: Deep
 Regionlets for Object Detection. arXiv preprint arXiv:1712.02408

\bibitem{szegedy2017inception}
Szegedy, C., Ioffe, S., Vanhoucke, V., Alemi, A.A.: Inception-v4,
  inception-resnet and the impact of residual connections on learning. In:
  AAAI. vol.~4, p.~12 (2017)

\bibitem{xie2017aggregated}
S.~Xie, R.~Girshick, P.~Doll{\'a}r, Z.~Tu, and K.~He.
\newblock Aggregated residual transformations for deep neural networks.
\newblock In {\em CVPR}, 2017.

\bibitem{zitnick2014edge}
Zitnick, C.L., Doll{\'a}r, P.: Edge boxes: Locating object proposals from
  edges. In: European Conference on Computer Vision. pp. 391--405. Springer
  (2014)

\bibitem{Uijlings2013Selective}
Uijlings, J.R., van~de Sande, K.E., Gevers, T., Smeulders, A.W.: Selective
  search for object recognition. International journal of computer vision
  104(2),  154--171 (2013)

\bibitem{chu2016structured}
Chu, Xiao and Ouyang, Wanli and Li, Hongsheng and Wang, Xiaogang: Structured
 feature learning for pose estimation. In: Proceedings of the IEEE Conference
 on Computer Vision and Pattern Recognition, 4715--4723 (2016)

\bibitem{everingham2015pascal}
Everingham, M., Eslami, S.A., Van~Gool, L., Williams, C.K., Winn, J.,
  Zisserman, A.: The pascal visual object classes challenge: A retrospective.
  International journal of computer vision  111(1),  98--136 (2015)

\bibitem{DBLP:journals/corr/abs-1711-08189}
Singh, B., Davis, L.S.: An analysis of scale invariance in object
  detection-snip. arXiv preprint arXiv:1711.08189  (2017)

\bibitem{russakovsky2015imagenet}
Deng, J., Dong, W., Socher, R., Li, L.J., Li, K., Fei-Fei, L.: Imagenet: A
  large-scale hierarchical image database. In: Computer Vision and Pattern
  Recognition, 2009. CVPR 2009. IEEE Conference on. pp. 248--255. IEEE (2009)

\bibitem{HeZR015}
Kaiming He and Xiangyu Zhang and Shaoqing Ren and Jian Sun.
\newblock Delving Deep into Rectifiers: Surpassing Human-Level Performance on ImageNet Classification.
\newblock {\em CoRR}, abs/1502.01852, 2015.

\bibitem{ioffe2015batch}
Ioffe, S., Szegedy, C.: Batch normalization: Accelerating deep network training
  by reducing internal covariate shift. In: International conference on machine
  learning. pp. 448--456 (2015)

\bibitem{goyal2017accurate}
Goyal, Priya and Doll{\'a}r, Piotr and Girshick, Ross and Noordhuis, Pieter and Wesolowski, Lukasz and Kyrola, Aapo and Tulloch, Andrew and Jia, Yangqing and He, Kaiming: 
Accurate, large minibatch SGD: training imagenet in 1 hour. arXiv preprint arXiv:1706.02677 (2017)

\bibitem{peng2017megdet}
Peng, Chao and Xiao, Tete and Li, Zeming and Jiang, Yuning and Zhang, Xiangyu and Jia, Kai and Yu, Gang and Sun, Jian:
 Megdet: A large mini-batch object detector. arXiv preprint arXiv:1711.07240 (2017)

\end{thebibliography}
}

\end{document}